\pdfoutput=1

\documentclass[11pt]{article}

\usepackage{authblk}

\usepackage[final]{ACL2023}

\usepackage{times}
\usepackage{latexsym}

\usepackage[T1]{fontenc}

\usepackage[utf8]{inputenc}

\usepackage{microtype}

\usepackage{inconsolata}

\usepackage{hyperref}
\usepackage{xstring}

\usepackage{color}

\usepackage{bm}
\usepackage{mathtools}
\usepackage{expex}
\usepackage{fontawesome}
\usepackage{longtable}
\usepackage{hyperref}

\title{A Decomposition-Based Approach for Evaluating and Analyzing Inter-Annotator Disagreement}

\author[1]{\textbf{Effi Levi}}
\author[1]{\textbf{Ari Rappoport}\protect\textsuperscript{\protect\hyperlink{ftnt_fn}{\dag}}}
\author[2]{\textbf{Shaul R. Shenhav}}
\affil[1]{Department of Computer Science, The Hebrew University of Jerusalem}
\affil[2]{Department of Political Science, The Hebrew University of Jerusalem}
\affil[ ]{{\tt efle@cs.huji.ac.il}} 
\affil[ ]{{\tt shaul.shenhav@mail.huji.ac.il}}

\begin{document}
\maketitle
\begingroup
\renewcommand{\thefootnote}{\dag}
\footnotetext{\hypertarget{ftnt_fn}{}Deceased}
\endgroup

\begin{abstract}
We propose a novel method to conceptually decompose an existing annotation into separate levels, allowing the analysis of inter-annotators disagreement in each level separately. We suggest two distinct strategies in order to actualize this approach: a \textit{theoretically-driven} one, in which the researcher defines a decomposition based on prior knowledge of the annotation task, and an \textit{exploration-based} one, in which many possible decompositions are inductively computed and presented to the researcher for interpretation and evaluation. Utilizing a recently constructed dataset for narrative analysis as our use-case, we apply each of the two strategies to demonstrate the potential of our approach in testing hypotheses regarding the sources of annotation disagreements, as well as revealing latent structures and relations within the annotation task. We conclude by suggesting how to extend and generalize our approach, as well as use it for other purposes.

\end{abstract}

\section{Introduction}
Annotated language resources play a fundamental role in both theoretical and computational linguistic research, aiding in analyzing and studying linguistic phenomena, as well as providing a ``ground truth'' or ``gold label'' for training supervised models for various NLP tasks. The annotation process commonly involves acquiring multiple annotator judgments on each of the data samples, a practice which is considered to increase annotation quality~\cite{snow2008cheap}. Inter-annotator disagreement may originate from various sources~\cite{artstein2017inter,beck-etal-2020-representation,basile2021we},
and is generally resolved using methods such as majority voting, averaging or resorting to expert judgment. While this may be perceived as relatively simple for ``objective'' tasks (e.g. part-of-speech tagging, syntactic parsing), more ``subjective'' 
tasks, where it is not always clear whether a ``correct'' judgment actually exists, may incur high levels of inter-annotator disagreement and pose a considerable challenge~\cite{alm2011subjective}. In such cases, it may be crucial to analyze and understand the sources for disagreement between the different annotators~\citep{bayerl-paul-2007-squibs,jiang2022investigating,sandri2023don}.

With that in mind, we propose to conceptually decompose the annotation task into separate sub-tasks or levels. This would allow studying inter-annotator disagreement separately for different components of the annotation (potentially containing varying degrees of disagreement), and to better understand the sources for the disagreement. 
In principle, this decomposition may be 
applied 
\textit{prior} to the annotation process. The annotation scheme could be decomposed to create a pipeline annotation procedure, therefore potentially reducing inter-annotator disagreement during the annotation process itself. This approach, however, has distinct shortcomings. Finding an effective decomposition may prove to be difficult, especially for highly complex annotation schemes (e.g., containing a large number of annotation possibilities per sample), and in many cases, the sources of disagreement may only reveal themselves during the annotation process. More than one decomposition may be of interest, and it is usually infeasible to repeat the annotation process for each such decomposition. In addition, allowing preconceptions to change the annotation process may influence the end result and create biases in the annotated dataset.

An alternative approach, and the one we take in this work, is applying the decomposition \textit{post-annotation}. This allows us to leverage the annotated dataset itself in order to detect and explore potential sources for disagreement in different aspects of the annotation scheme. We suggest two different strategies to actualize this approach: a \textit{theoretically-driven} one, in which the researcher, based on her knowledge of the annotation scheme
and theoretical expectations,
defines 
a theoretically meaningful 
decomposition according to expected sources of subjectivity in the annotation scheme. This may be an effective way for the researcher to test hypotheses regarding the sources of disagreement in the annotated dataset. 
While this strategy may only be used to confirm or refute existing hypotheses, latent sources of disagreement may still exist.
A second strategy is an \textit{exploration-driven} one, in which many possible decompositions are inductively computed and presented to the researcher for interpretation and evaluation. This strategy could potentially reveal latent structures and relations in the annotation scheme that may have led to the inter-annotator disagreements. 

In order to test and explore our approach, we direct our attention to the task of \textit{narrative analysis}, which is highly subjective by nature. Several works have constructed and annotated datasets for capturing various narrative aspects in texts~\cite{baiamonte2016annotating,delmonte-marchesini-2017-semantically,papalampidi2019movie,volpetti2020temporal,swanson-etal-2014-identifying,li2017annotating,saldias2020exploring}. As our use-case, we choose to utilize the recent work by~\citet{levi2022}, who have developed a sentence-level multi-label annotation scheme, consisting of three different narrative plot elements: \textit{Complication}, \textit{Resolution} and \textit{Success}, and used it to annotate a dataset of news article sentences. Together, these three elements capture a typical tension-release plot type, varied by the different possible elements combinations.

First, we apply the \textit{theoretically-driven} strategy to perform a conceptual decomposition of the annotation task into two separate questions/levels,
each addressing a different level of disagreement: 
(1) \textbf{whether} or not a narrative plot exists in the text, and (2) \textbf{which} plot elements exist in the text. We then employ statistical analysis in order to quantify how much of the inter-annotator disagreement can be explained by each of the two levels. We make two main observations: first -- there is notable disagreement on the first level, i.e., on whether or not there is a narrative plot in the sentence at all; and second -- given an agreement on the first level, there is a notable improvement in the agreement on the second level, i.e., on which of the narrative elements appear in the sentence. In addition, we briefly discuss possible sources for disagreement in each of the two decomposition levels.

Next, applying the \textit{exploration-driven} strategy, we compute all the possible decompositions over the annotation scheme, and for each of them, we perform the same statistical analysis to produce the decomposed inter-annotator agreement scores. We then use two different criteria to rank these decompositions: based on first-level agreement, and based on the agreement on the three narrative elements, given an agreement on the first level (second-level agreement).
We present and discuss several interesting decompositions which ranked high according to each of the criteria. 

We conclude by suggesting several promising directions for future work: extending our approach to include more levels of decomposition, as well as decompositions into more than two sets; generalizing our approach to different types of annotation tasks; and utilizing it to improve training of supervised models.

\section{Experimental Setup}
\label{sec:setup}

\subsection{Data}
\label{subsec:data}
We used the NEAT dataset, created by \newcite{levi2022} for narrative analysis, consisting of 2,209 sentences (taken from news articles). The dataset was annotated according to a sentence-level multi-label narrative annotation scheme. The authors employed a total of three annotators (henceforth denoted as $A_1$, $A_2$ and $A_3$).\footnote{For the purpose of our analysis, we used only the annotations made by the three annotators, and discarded the ``gold labels'' which were obtained by a third expert annotator.} 
The annotators assigned each sentence with a subset of the three narrative elements \textit{Complication}, \textit{Resolution} and \textit{Success}. These elements are not mutually exclusive, so a sentence may be annotated with any combination of the three, or none at all. All the sentences were annotated by $A_1$, while each sentence was additionally annotated by either $A_2$ or $A_3$ (statistics are given in Table~\ref{tab:annotators}). Further details on the annotation scheme, guidelines and dataset can be found in~\newcite{levi2022} and in \url{https://github.com/efle/neat}.

\begin{table}[!htbp]
\centering
\begin{tabular}{ccccc}
    \hline
    & \textbf{\# Sentences} & \textbf{Comp.} & \textbf{Res.} & \textbf{Suc.} \\
    \hline
    $A_1$ & 2,209 & 1,083 & 448 & 312 \\
    $A_2$ & 1,135 & 504 & 262 & 137 \\
    $A_3$ & 1,074 & 569 & 296 & 168 \\
    \hline
\end{tabular}
\caption{Annotation statistics: the number of annotated sentences (out of a total of 2,209) and the number of sentences annotated with each of the three narrative elements (\textit{Complication}, \textit{Resolution} and \textit{Success}), for each of the three annotators.
}
\label{tab:annotators}
\end{table}

\subsection{Agreement Metrics}

In order to measure inter-annotator agreement, we utilize two commonly-used metrics. The first is the pairwise percent agreement (PPA), which is the percentage of instances on which the annotators agreed (i.e. assigned identical annotation) out of all the instances which were annotated by both. While this is the simplest and most intuitive way to measure agreement, it has the drawback of not accounting for chance agreement. This may pose a problem in some cases, e.g. in cases of significant class imbalance, as is our case (evident in Table~\ref{tab:annotators}). The second metric we employ, Cohen's Kappa coefficient ($\kappa$), was designed to address this very issue~\cite{cohen1960coefficient,landis1977measurement} by correcting the PPA for chance agreement between the annotators. It is defined as:

\begin{equation}
    \kappa = \frac{p_o - p_e}{1 - p_e}
\end{equation}

where $p_o$ is the observed agreement (namely, the PPA), and $p_e$ is the expected probability for chance agreement, which is calculated from the individual empirical distributions of the annotators over the annotated instances.

Table~\ref{tab:agreement} summarizes the overall inter-annotator agreement for the NEAT dataset. For each pair of annotators, we report PPA \& $\kappa$ on each of the three narrative elements in the annotation scheme. We observe no significant differences between the two pairs in terms of per-element agreement (a max. difference of $1.8\%$ in PPA and $0.04$ in $\kappa$). 
It is, however, interesting to note the different pictures painted by PPA vs. $\kappa$. Among the three narrative elements, the highest average PPA ($92.8\%$) is achieved by \textit{Success}, while the highest average $\kappa$ (0.79) is achieved by \textit{Complication}, which is three time more frequent (see Table~\ref{tab:annotators}). This clearly demonstrates the $\kappa$ coefficient's advantage in compensating for class imbalance. Consequently, in the remainder of this paper we shall use $\kappa$ to measure inter-annotator agreement.

\begin{table}[!htbp]
\begin{center}
\begin{tabular}{c|ccc|c}
    \hline
    \textbf{PPA} & \textit{Comp.} & \textit{Res.} & \textit{Suc.} & Average\\
    \hline
    $A_1A_2$ & 88.7\% & 88.2\% & 92.9\% & 89.9\% \\
    $A_1A_3$ & 90.5\% & 86.8\% & 92.7\% & 90.0\% \\
    \textbf{Average} & \textbf{89.6\%} & \textbf{87.5\%} & \textbf{92.8\%} & \textbf{90.0\%} \\
    \hline
    $\bm{\kappa}$ & \textit{Comp.} & \textit{Res.} & \textit{Suc.} & Average\\
    \hline
    $A_1A_2$ & 0.77 & 0.65 & 0.68 & 0.70 \\
    $A_1A_3$ & 0.81 & 0.64 & 0.72 & 0.72 \\
    \textbf{Average} & \textbf{0.79} & \textbf{0.65} & \textbf{0.70} & \textbf{0.71} \\
    \hline
\end{tabular}
\caption{Overall inter-annotator agreement}
\label{tab:agreement}
\end{center}
\end{table}

\section{Theoretically-Driven Decomposition}
\label{sec:theo_decomposition}

\subsection{Process}
\label{subsec:decomposition}

The concept of ``narrative plot'' is an elusive one. Even though there is general agreement among narratologists that succession of events is the core of narrativity~\cite{abbott2021cambridge,genette1980narrative,rimmon2003narrative}, 
determining whether a plot is present at all in a given text is still a challenging task~\cite{shenhav2015analyzing}. 
This fundamental question guided us in defining a theoretically-based decomposition for the NEAT annotation task. The three narrative elements -- \textit{Complication}, \textit{Resolution}, and \textit{Success} -- capture a typical tension-release plot type, varied by the different possible element combinations. Consequently, we decompose the annotation into two separate levels: \textbf{whether} a narrative plot was identified in the sentence, and \textbf{which} narrative elements were detected in the sentence. 
To do that, we define a new label denoted ``\textit{Plot}'', which represents the existence of a narrative plot as defined by our three narrative elements. From a theoretical perspective, this label indicates whether or not the sentence facilitates the plot type captured by this annotation scheme. We compute this label from the existing annotations as follows. Given a sentence $s$ annotated by annotator $A_i$ ($i \in \{1, 2, 3\}$), and a narrative element $e \in E = \{\textit{Complication}, \textit{Resolution}, \textit{Success}\}$, let us define:

\begin{equation}
    A^e_i(s) = 
    \begin{dcases} 
      1 & \text{$A_i$ detected element $e$ in $s$} \\
      0 & \text{Otherwise}
    \end{dcases}
\end{equation}

Then

\begin{equation}
    A^\textit{Plot}_i(s) = \bigvee_{e \in E}{A^e_i(s)}
\end{equation}

Hence, \textit{Plot} is considered to be identified in a sentence by an annotator if and only if the annotator detected at least one of the three narrative elements in that sentence. 
Next, we quantify how much of the inter-annotator disagreement is explained by each of the two separate levels of the annotation. For this purpose, we perform the following 2-stage procedure for each pair of annotators -- $A_1A_2$ and $A_1A_3$:
\begin{enumerate}
    \item Measure the inter-annotator agreement for the \textit{Plot} label, to obtain the agreement for the first level of the annotation.
    \item Neutralize the disagreement stemming from the first level by discarding all the sentences on which the annotators disagree on the \textit{Plot} label, leaving only the sentences on which they agree. By measuring the agreement on the remaining sentences, we obtain the inter-annotator agreement for the second level of the annotation.
\end{enumerate}

\subsection{Analysis \& Discussion} 
\label{subsec:analysis}

Table~\ref{tab:agreement_factored} summarizes the inter-annotator agreement factored into the two decomposed levels of the narrative annotation -- \textbf{whether} and \textbf{which} -- following the procedure described in Section~\ref{subsec:decomposition}. 
For each pair of annotators we report $\kappa$ on the \textit{Plot} label (which represents the first level of the annotation), as well as for the three narrative elements given an agreement on the \textit{Plot} label (representing the second level of the annotation). Differences between the two pairs, in terms of per-element agreement, remain largely insignificant (a max. difference of $0.04$).

\begin{table}[!htbp]
\begin{center}
\resizebox{\columnwidth}{!}{%
\begin{tabular}{c|c|cccc}
    \hline
    & \textit{Plot} & \textit{Comp.} & \textit{Res.} & \textit{Suc.} & Average \\
    \hline
    $A_1A_2$ & 0.71 & 0.91 & 0.77 & 0.80 & 0.83 \\
    $A_1A_3$ & 0.69 & 0.91 & 0.73 & 0.84 & 0.83\\
    \textbf{Average} & \textbf{0.70} & \textbf{0.91} & \textbf{0.75} & \textbf{0.82} & \textbf{0.83} \\
    \hline
\end{tabular}
}
\caption{Inter-annotator agreement, factored into the two decomposed annotation levels: \textbf{whether} the sentence contains a narrative plot (``\textit{Plot}''), and \textbf{which} narrative elements the sentence contains, given an agreement on \textit{Plot} (``\textit{Comp.}'', ``\textit{Res.}'' and ``\textit{Suc.}'')}
\label{tab:agreement_factored}
 \end{center}
\end{table}

\subsubsection{First Level Agreement}

We observe a notable disagreement between the annotators on the first level (i.e., the ``\textit{Plot}'' label), indicated by an average $\kappa$ of 0.70. This implies that even before the distinction between the different narrative elements, there is a fundamental difference between the annotators in terms of their interpretation of the concept of narrative plot. 

There are several possible sources for the inter-annotator disagreement on the ``\textit{Plot}'' label. For example, consider the following sentence:
\ex<whetherSchemeReferential>
Trump on the DNC hack during the second presidential debate ($A_1$: \textit{None}, $A_2$: \textit{Complication})
\xe
The sentence contains a reference to a complicative plot element -- ``the DNC hack'' -- which is not directly described or discussed in the sentence itself, creating a narrative gap which the reader should complete in order infer this complicative plot development. This ``referential narrative'' is a rhetorical device which is commonly used in informational text; therefore, this type of disagreement may be resolved by formally defining how to handle such cases in the scheme level (essentially not leaving the decision to the annotators).

However, disagreement is sometimes rooted in the interpretation of a specific phrasing rather than in generalizable narrative aspects. For example,
\ex
A 1984 rebranding included the brand's targeting the high-end luxury market. ($A_1$: \textit{None}, $A_2$: \textit{Resolution})
\xe
In this case, it seems that the disagreement between the annotators stems from the term ``rebranding'' in relation to the brand’s targeting of the high-end luxury market. $A_2$ perceives this action as resolving a situation (or an existing tension), while $A_1$ does not. The annotators 
seem to 
disagree on the \textit{narrative interpretation} of 
the text. Unlike example (\getref{whetherSchemeReferential}),
this disagreement is hard to resolve via a formal definition in the scheme level, since it involves the subjective interpretation of a specific word (``branding'') within the context of a particular sentence. 

\subsubsection{Second Level Agreement}

Agreement on the second level (i.e., the agreement scores on the three narrative elements, after neutralizing the first level) is significantly higher than the overall agreement reported in Table~\ref{tab:agreement}, with an average increase of 0.12. This suggests that once the first level is resolved (i.e., it is agreed that the text contains a narrative plot), 
reaching agreement on the second level is significantly easier
(i.e., the specific elements of which the plot consists). 

As in the first level, second-level annotation disagreements originate from various sources. For example:
\ex<whichSchemeNegation>
``History has suggested that the pessimists have been wrong time and time again,'' he said. ($A_1$: \textit{Complication}, $A_3$: \textit{Resolution})
\xe
This sentence stresses the difference between the complication-resolution plot structure and the negative-positive polarity. $A_1$ detects a complicating function in ``have been wrong time and time again''; however, $A_3$
views the fact that it were the ``pessimists'' who have been wrong as a double negation of a sort and thus as a resolved situation, correlating the complication-resolution structure with the negative-positive polarity. 

Disagreement may also stem from text ambiguity on the semantic level:
\ex<whichTextAmbiguity1>
This bleeding is not expected, at least in such a significant population so quickly. ($A_1$: \textit{Complication}, $A_2$: \textit{Resolution})
\xe
The origin of disagreement is the ambiguous start of the sentence -- ``This bleeding is not expected''. This can be understood in two ways: either the bleeding was not expected but has happened, in which case a complication has transpired ($A_1$), or the bleeding is not expected to happen, in which case a resolution has been reached ($A_2$). This is a noteworthy example as to how text ambiguity may produce two possible narrative interpretations.

\section{Exploring Other Decompositions}
\label{sec:exp_decomposition}

In Section~\ref{subsec:decomposition} we described a decomposition of the annotation according to a specific theoretical reasoning (grounded in the annotation scheme). However, other possible decompositions exist. Using different decompositions to perform the analysis described in Section~\ref{subsec:decomposition} may reveal additional latent conceptual structures in the annotation, potentially providing new insights into the sources of the inter-annotator disagreement. Here, we describe a method to explore such decompositions, apply it to the NEAT dataset, and discuss the results.

\subsection{Process}
\label{subsec:full_dec_process}

In order to facilitate this discussion, we will represent each possible label combination in NEAT as a 3-bit sequence, respectively corresponding with \textit{Complication}, \textit{Resolution} and \textit{Success}. For example, the sequence $101$ represents the combination \textit{Complication} \& \textit{Success}, while the sequence $010$ represents just a \textit{Resolution}. The complete encoding is given in Table~\ref{tab:combo_encoding}.

\begin{table}[!htbp]
\centering
\begin{tabular}{c|ccc}
    \hline
    \textbf{Encoding} & \textit{Comp.} & \textit{Res.} & \textit{Suc.} \\
    \hline
    000 & \faClose & \faClose & \faClose \\
    001 & \faClose & \faClose & \faCheck \\
    010 & \faClose & \faCheck & \faClose \\
    011 & \faClose & \faCheck & \faCheck \\
    100 & \faCheck & \faClose & \faClose \\
    101 & \faCheck & \faClose & \faCheck \\
    110 & \faCheck & \faCheck & \faClose \\
    111 & \faCheck & \faCheck & \faCheck \\
    \hline
\end{tabular}
\caption{3-bit Encoding for all the possible label combinations in NEAT}
\label{tab:combo_encoding}
\end{table}

We define a \textit{decomposition} of the annotation scheme as a partition of the set of eight possible label combinations $S=\{000,$ $001,$ $011,$ $100,$ $101,$ $110,$ $111\}$ into two disjoint sets $S_1 \in S$,  $S_2 \in S$ such that $S_1 \cup S_2 = S$. This partition defines a new binary label over the dataset -- denoted $C_{S_1}$ -- where every sentence labeled with one of the label combinations in $S_1$ is labeled with $0$, and every sentence labeled with one of the label combinations in $S_2$ is labeled with $1$. $S$ contains $2^8=256$ possible subsets; however, since every partition $(S_1, S_2)$ is equivalent to $(S_2, S_1)$, there are $\frac{256}{2}=128$ possible partitions of $S$ into two disjoint sets. Discarding the null partition $(\{\}, S)$, we arrive at 127 possible decompositions for $S$.

For example, the decomposition $S_1=\{100, 010, 001\}, S_2=\{000, 011, 101, 110, 111\}$ defines a new binary label $C_{\{100, 010, 001\}}$ where every sentence containing exactly one of the three narrative elements (\textit{Complication}, \textit{Resolution} and \textit{Success}) is labeled with $0$, and all other sentences are labeled with $1$. As another example, the decomposition defined by $S_1=\{000\}$ corresponds with the one described in Section~\ref{subsec:decomposition}: a new binary label ($C_{\{000\}}$), in which every sentence with no narrative element ($000$) is labeled with $0$, while every sentence that contains any narrative element ($001, 011, 100, 101, 110, 111$) is labeled with $1$.

\begin{table*}[!htbp]
\centering
\resizebox{\textwidth}{!}{%
\begin{tabular}{cc|c|ccc|c}
    \hline
    & & & \multicolumn{4}{c}{\textbf{2\textsuperscript{nd} Level}} \\
    $\bm{S_1}$ & $\bm{S_2}$ & \textbf{1\textsuperscript{st} Level} &  \textit{Complication} & \textit{Resolution} & \textit{Success} & \textbf{Average}\\
    \hline
    101 & 000,001,010,011,100,110,111 & 0.35 & 0.80 & 0.66 & 0.72 & 0.72 \\
    010,101 & 000,001,011,100,110,111 & 0.42 & 0.83 & 0.74 & 0.76 & 0.78 \\
    010 & 000,001,011,100,101,110,111 & 0.45 & 0.83 & 0.73 & 0.74 & 0.77 \\
    010,111 & 000,001,011,100,101,110 & 0.48 & 0.84 & 0.74 & 0.76 & 0.78 \\
    010,101,111 & 000,001,011,100,110 & 0.48 & 0.84 & 0.73 & 0.80 & 0.79\\
    001,010,101 & 000,011,100,110,111 & 0.51 & 0.83 & 0.77 & 0.87 & 0.82 \\
    001,010 & 000,011,100,101,110,111 & 0.51 & 0.84 & 0.76 & 0.84 & 0.81 \\
    011,101 & 000,001,010,100,110,111 & 0.52 & 0.81 & 0.68 & 0.76 & 0.75 \\
    101,110 & 000,001,010,011,100,111 & 0.52 & 0.83 & 0.75 & 0.75 & 0.78 \\
    001,010,111 & 000,011,100,101,110 & 0.53 & 0.83 & 0.76 & 0.88 & 0.82 \\
    \hline
\end{tabular}
}
\caption{Top ten decompositions with the lowest first-level inter-annotator agreement scores. $\bm{S_1}$ and $\bm{S_2}$: the decomposition; \textbf{1\textsuperscript{st} Level}: inter-annotator agreement ($\kappa$) on $C_{S_1}$; \textbf{Complication}, \textbf{Resolution} and \textbf{Success}: inter-annotator agreement ($\kappa$) on each element given an agreement on $C_{S_1}$;  \textbf{Average}: second-level inter-annotator agreement, averaged over the three narrative elements. All the statistics were averaged over the two annotator pairs.}
\label{tab:exp_1st_level}
\end{table*}

Given a decomposition $S_1, S_2$, we perform the same 2-step procedure described in Section~\ref{subsec:decomposition} to calculate the decomposed agreement scores:
\begin{enumerate}
    \item Measure the inter-annotator agreement for $C_{S_1}$, to obtain the agreement for the first level of the annotation.
    \item Neutralize the disagreement stemming from the first level by discarding all the sentences on which the annotators disagree on $C_{S_1}$, leaving only the sentences on which they agree. By measuring the agreement on the remaining sentences, we obtain the inter-annotator agreement for the second level of the annotation.
\end{enumerate}

\subsection{Analysis \& Discussion}

For each possible decomposition, we performed the above-mentioned 2-step procedure to receive the decomposed $\kappa$ agreement scores. A full summary of the results can be found in Appendix~\ref{app:full_results}. We use two different criteria to rank these decompositions: based on the agreement level on $C_{S_1}$ (first-level agreement), and based on the average agreement over the three narrative elements, given an agreement over $C_{S_1}$ (second-level agreement). We present and discuss several decompositions which ranked high according to each of the criteria. 

\subsubsection{First Level Agreement}

As the first criterion, we used the inter-annotator agreement on $C_{S_1}$ averaged over the two annotator pairs to rank all the decompositions, lowest to highest. Table~\ref{tab:exp_1st_level} lists the ten decompositions with the lowest first-level agreement scores. For each decomposition, the table lists:

\begin{itemize}
    \item $\bm{S_1}$ and $\bm{S_2}$: the decomposition
    \item \textbf{1\textsuperscript{st} Level}: inter-annotator agreement ($\kappa$) on the first level of the annotation ($C_{S_1}$), averaged over the two annotator pairs
    \item \textbf{Complication}, \textbf{Resolution} and \textbf{Success}: inter-annotator agreement ($\kappa$) on each of the three narrative elements in NEAT given an agreement on $C_{S_1}$, each averaged over the two annotator pairs
    \item \textbf{Average}: second-level inter-annotator agreement, averaged over the three narrative elements
\end{itemize}

Some of these decompositions present interesting cases for analysis and discussion. For example, the first decomposition in the table is $S_1=\{101\}$, $S_2=\{000,001,010,011,100,110,111\}$, has a first-level inter-annotator agreement score of $\kappa=0.35$. It represents a division of the annotation into sentences in which a \textit{Complication} is accompanied by a \textit{Success} without a \textit{Resolution}, vs. all other sentences. For example,
\ex
I think I lost the race to a great competitor. ($A_1$: \textit{Complication, Success}, $A_2$: \textit{Complication})
\xe
From a narrative perspective, this specific combination ($101$) is very unique, as it is very unlikely to encounter a \textit{Complication} with an ensuing \textit{Success} without an inter-connecting \textit{Resolution}. This uniqueness is also evident in the annotated dataset: only 1\% of the sentences were annotated with this label combination by $A_1$, 0.6\% by $A_2$ and 0.7\% by $A_3$. It stands to reason that such a unique and unlikely label combination would produce high inter-annotator disagreement. We can also observe that second-level agreement is largely unaffected compared to the overall agreement (presented in Table~\ref{tab:agreement}), due to the very low number of occurrences in the dataset.

\begin{table*}[!htbp]
\centering
\resizebox{\textwidth}{!}{
\begin{tabular}{cc|c|ccc|c}
    \hline
    & &  & \multicolumn{4}{c}{\textbf{2\textsuperscript{nd} Level}} \\
    $\bm{S_1}$ & $\bm{S_2}$ & \textbf{1\textsuperscript{st} Level} &  \textit{Complication} & \textit{Resolution} & \textit{Success} & \textbf{Average}\\
    \hline
    000,011,101,110 & 001,010,100,111 & 0.55 & 0.94 & 0.88 & 0.90 & 0.91 \\
    001,010,100 & 000,011,101,110,111 & 0.57 & 0.94 & 0.89 & 0.87 & 0.90 \\
    000,101,110 & 001,010,011,100,111 & 0.57 & 0.94 & 0.85 & 0.89 & 0.89 \\
    000,010,101,110 & 001,011,100,111 & 0.64 & 0.93 & 0.79 & 0.97 & 0.89 \\
    000,011,110,111 & 001,010,100,101 & 0.57 & 0.94 & 0.90 & 0.85 & 0.89 \\
    000,011,110 & 001,010,100,101,111 & 0.57 & 0.93 & 0.87 & 0.88 & 0.89 \\
    000,011,100,111 & 001,010,101,110 & 0.54 & 0.81 & 0.95 & 0.92 & 0.89 \\
    000,101,110,111 & 001,010,011,100 & 0.59 & 0.95 & 0.86 & 0.87 & 0.89 \\
    000,110 & 001,010,011,100,101,111 & 0.60 & 0.93 & 0.85 & 0.88 & 0.89 \\
    000,110,111 & 001,010,011,100,101 & 0.59 & 0.94 & 0.88 & 0.84 & 0.88 \\
        \hline
\end{tabular}
}
\caption{Top ten decompositions with highest second-level inter-annotator agreement scores. $\bm{S_1}$ and $\bm{S_2}$: the decomposition; \textbf{1\textsuperscript{st} Level}: inter-annotator agreement ($\kappa$) on $C_{S_1}$; \textbf{Complication}, \textbf{Resolution} and \textbf{Success}: inter-annotator agreement ($\kappa$) on each element given an agreement on $C_{S_1}$;  \textbf{Average}: second-level inter-annotator agreement, averaged over the three narrative elements. All the statistics were averaged over the two annotator pairs.}
\label{tab:exp_2nd_level}
\end{table*}

Another high-ranking interesting case is the decomposition $S_1=\{010\}$, $S_2=\{000$, $001$, $011$, $100$, $101$, $110$ ,$111\}$, with a first-level inter-annotator agreement score of $\kappa=0.45$. This decomposition represents a division of the annotation to sentences which exclusively contain a \textit{Resolution} (breaking the conventional coherent logic of the complication-resolution sequence~\cite{labov1967narrative}), vs. all other sentences. We hypothesize that a \textit{Resolution} is harder to define by itself, without the context of a \textit{Complication} leading to it or a \textit{Success} to ground and validate it, which may explain the low first-level agreement in this case. For example, the sentence
\ex 
"We're going to have to do things that we never did before," he told Yahoo News. ($A_1$: \text{None}, $A_2$: \textit{Resolution})
\xe
contains no complicating narrative nor an explicit success, making it harder to decide whether or not it describes an actual resolving action. Unlike the previous decomposition, in this case the second-level agreement is higher than the overall agreement (0.77 vs. 0.70), meaning that theoretically, were we able to find a way to resolve the first-level disagreement, we would get a notable improvement in the agreement over the three narrative elements.

A third interesting case is the decomposition $S_1=\{001, 010\}$, $S_2=\{000$, $011$, $100$, $101$, $110$ ,$111\}$, with a first-level inter-annotator agreement of $\kappa=0.51$. This decomposition discriminates between sentences annotated with either only a \textit{Resolution} or only a \textit{Success}, and sentences annotated with any other label combination. 
Interestingly, from a narrative perspective, the label combinations set $S_1=\{001, 010\}$ captures somewhat optimistic-driven plots. However, it is important to note that it does not include the label combination 011 (both a \textit{Resolution} and a \textit{Success}), 
meaning that It captures optimism as a state, rather than as the outcome of a narrative process in which, for example, a \textit{Complication} is resolved or a \textit{Resolution} develops into \textit{Success};
this might explain the low first-level agreement, as \textit{Resolution} and \textit{Success} are somewhat correlated, which may make it harder to define cases in which only one of them is present. Note that in this case, there is a significant difference between the second-level agreement and the overall agreement (0.81 vs. 0.70), meaning that if we do manage to resolve the disagreement on the first level of the decomposition, we will be awarded with a significant increase in agreement over the three narrative elements.

\subsubsection{Second Level Agreement}
\label{subsubsec:exp_2nd_level}

For a second criterion, we computed (for each decomposition) the average inter-annotator agreement on the second level -- the three narrative elements -- given an agreement on the first level ($C_{S_1}$). We averaged this score over the two annotator pairs, and used it to re-rank all the decompositions, highest to lowest. Table~\ref{tab:exp_2nd_level} lists the ten decompositions with the highest average second-level agreement.

Comparing Table~\ref{tab:exp_1st_level} and Table~\ref{tab:exp_2nd_level}, we can observe that high second-level agreement decompositions generally seem more balanced than low first-level agreement ones (in terms of the number of combinations in $S_1$ and $S_2$).
However, large and diverse sets of label combinations are harder to find a theoretical explanation for (in terms of the original annotation scheme). This could potentially mean that high second-level agreement decompositions are generally less interpretable.

One decomposition in Table~\ref{tab:exp_2nd_level} that does stand out in terms of interpretability is $S_1=\{001, 010, 100\}, S_2=\{000, 011, 101, 110, 111\}$. This decomposition represents a division of the annotation into cases that contain \textit{exactly} one narrative element vs. all other cases.  If we think of the number of narrative elements in the sentence as its ``narrative cardinality'', then this decomposition discriminates sentences with ``narrative cardinality'' of 1 from sentences with other levels of ``narrative cardinality'' (0, 2 or 3). Evidently, decomposing the annotation based on ``narrative cardinality'' results in high second-level agreement, as this specific decomposition achieves an averaged second-level agreement $\kappa$ of 0.90 (the second-highest among all possible decompositions). Fully decomposing the annotation according to ``narrative cardinality'' would require decomposition into the four possible ``narrative cardinalities'' (0, 1, 2 and 3), making decomposition into more than two sets a promising direction for future work.

\section{Conclusion}
\label{sec:conclusion}

We proposed a novel approach for exploring sources of inter-annotator disagreement by decomposing an existing annotation into separate components or levels. This approach allows the evaluation and analysis of the sources of inter-annotator disagreement separately for each level of the annotation. We suggested two distinct strategies to actualize our approach: \textit{theoretically-driven}, guided by prior knowledge of the annotation scheme, and \textit{exploration-driven}, in which many possible decompositions are inductively computed and considered for interpretation and evaluation.

As a use-case, we utilized a narrative analysis dataset, annotated with a specific tension-release plot type, composed of three narrative elements (\textit{Complication}, \textit{Resolution} and \textit{Success}) by two annotator pairs~\cite{levi2022}. First, we based the theoretically-driven strategy on the fundamental question of whether or not a narrative plot exists in the text, by decomposing the annotation task into two separate levels: (1) \textbf{whether} or not a narrative plot exists in the text, and (2) \textbf{which} plot elements exist in the text. We quantified how much of the inter-annotator disagreement is explained by each of the two levels, discovering a notable disagreement on the first level, as well as a significant improvement in the agreement on the second level given an agreement on the first level. We additionally demonstrated possible sources for disagreement in each of the two decomposition levels through several examples.

Next, we applied the \textit{exploration-driven} strategy by computing all the possible decompositions over the annotation scheme, and producing the decomposed inter-annotator agreement scores for each. We used two different criteria to rank these decompositions: based on first-level agreement, and based on the average agreement on the three narrative elements, given an agreement on the first level (second-level agreement). Based on these rankings, we identified and discussed several informative decompositions which revealed interesting insights into the annotation task.

The decomposition approach presented here may have general implications on the analysis of inter-annotator agreement. Given an annotated dataset, it allows conceptually decomposing the annotation by theoretically or empirically driven aspects of the scheme, and statistically quantifying the distribution of the disagreement over these chosen aspects. In a larger sense, it offers a more flexible approach than current conventions of inter-annotation diagnosis, allowing researchers to study and evaluate various aspects and sources of disagreement in annotation tasks.

There are several exciting directions for future work. The process can easily be applied to any multi-class or multi-label annotation scheme, but it may also be extended and generalized to apply to other types of annotation schemes, such as structured annotation tasks (sequence tagging or even deep parsing tasks). Other, more complex criteria may be devised in order to efficiently identify informative decompositions in the exploration-based strategy. Following the insight gained in Section~\ref{subsubsec:exp_2nd_level} on ``narrative cardinality'', it would be beneficial to examine decompositions into a larger number of sets (rather than just two), or explore the possibility of deeper decompositions, with more than two levels. Finally, our proposed approach may be employed to train better supervised models over annotated datasets. Most supervised models are trained on a single ``ground truth'', without taking into account inter-annotator disagreements.
Given a specific decomposition, we can train a series of pipelined supervised models to imitate the decomposition process. Using our exploration-based strategy, we can leverage the variability inherent in multiple annotations to discover decompositions that optimize this procedure and produce better supervised models for the task at hand.

\section*{Limitations}

The primary limitation of this work is the confinement of our empirical demonstration to a single annotated dataset. However, given the strict space constraints, our main objective was to introduce our method by providing a comprehensive deep dive into one specific use-case. This allowed us to demonstrate the variety of insights that may be produced by using the presented method to reveal latent structures in the analyzed annotation scheme and data, rather than providing a broader yet superficial analysis of several datasets. Focusing on a single use-case allowed us to provide a thorough demonstration of our method's capabilities. We specifically chose the NEAT annotation scheme due to its highly subjective nature, providing an opportunity to reveal various aspects of annotator disagreement. Nonetheless, establishing the generalizability of our method across additional annotation schemes and datasets remains an important avenue for future work.

\bibliography{arxiv_v3}
\bibliographystyle{acl_natbib}

\appendix

\section{Full Decomposition Results}
\label{app:full_results}

Table~\ref{tab:full_results} contains inter-annotator agreement, factored into two decomposed levels,  for each of the 127 possible decompositions of NEAT (the process is described in Section~\ref{subsec:full_dec_process}). For each decomposition, the table lists the following:

\begin{itemize}
    \item $\bm{S_1}$ and $\bm{S_2}$: the decomposition
    \item \textbf{1\textsuperscript{st} Level}: inter-annotator agreement ($\kappa$) on the first level of the annotation ($C_{S_1}$), averaged over the two annotator pairs
    \item \textbf{Complication}, \textbf{Resolution} and \textbf{Success}: inter-annotaotr agreement ($\kappa$) on each of the three narrative elements in NEAT given an agreement on $C_{S_1}$ (representing the second level of the annotation), each averaged over the two annotator pairs
    \item \textbf{Average}: second-level inter-annotator agreement, averaged over the three narrative elements
\end{itemize}

\onecolumn
{\small\tabcolsep=5pt
\begin{longtable}[c]{cc|c|ccc|c}
    \hline
    & & & \multicolumn{4}{c}{\textbf{2\textsuperscript{nd} Level}} \\
    $\bm{S_1}$ & $\bm{S_2}$ & \textbf{1\textsuperscript{st} Level} &  \textit{Complication} & \textit{Resolution} & \textit{Success} & \textbf{Avergae} \\
    \hline
    000 & 001,010,011,100,101,110,111 & 0.70 & 0.91 & 0.75 & 0.82 & 0.83 \\
001 & 000,010,011,100,101,110,111 & 0.60 & 0.80 & 0.68 & 0.81 & 0.76 \\
010 & 000,001,011,100,101,110,111 & 0.45 & 0.83 & 0.73 & 0.74 & 0.77 \\
011 & 000,001,010,100,101,110,111 & 0.57 & 0.80 & 0.67 & 0.75 & 0.74 \\
100 & 000,001,010,011,101,110,111 & 0.74 & 0.92 & 0.78 & 0.73 & 0.81 \\
101 & 000,001,010,011,100,110,111 & 0.35 & 0.80 & 0.66 & 0.72 & 0.73 \\
110 & 000,001,010,011,100,101,111 & 0.54 & 0.83 & 0.75 & 0.74 & 0.77 \\
111 & 000,001,010,011,100,101,110 & 0.58 & 0.80 & 0.65 & 0.72 & 0.72 \\
000,001 & 010,011,100,101,110,111 & 0.74 & 0.93 & 0.80 & 0.70 & 0.81 \\
000,010 & 001,011,100,101,110,111 & 0.72 & 0.97 & 0.66 & 0.88 & 0.84 \\
000,011 & 001,010,100,101,110,111 & 0.66 & 0.92 & 0.76 & 0.85 & 0.84 \\
000,100 & 001,010,011,101,110,111 & 0.67 & 0.81 & 0.93 & 0.87 & 0.87 \\
000,101 & 001,010,011,100,110,111 & 0.68 & 0.91 & 0.76 & 0.85 & 0.84 \\
000,110 & 001,010,011,100,101,111 & 0.60 & 0.93 & 0.85 & 0.88 & 0.89 \\
000,111 & 001,010,011,100,101,110 & 0.67 & 0.92 & 0.76 & 0.85 & 0.84 \\
001,010 & 000,011,100,101,110,111 & 0.51 & 0.84 & 0.76 & 0.84 & 0.81 \\
001,011 & 000,010,100,101,110,111 & 0.69 & 0.81 & 0.65 & 0.91 & 0.79 \\
001,100 & 000,010,011,101,110,111 & 0.66 & 0.93 & 0.83 & 0.83 & 0.86 \\
001,101 & 000,010,011,100,110,111 & 0.59 & 0.79 & 0.69 & 0.84 & 0.77 \\
001,110 & 000,010,011,100,101,111 & 0.53 & 0.84 & 0.80 & 0.86 & 0.83 \\
001,111 & 000,010,011,100,101,110 & 0.61 & 0.80 & 0.68 & 0.86 & 0.78 \\
010,011 & 000,001,100,101,110,111 & 0.57 & 0.84 & 0.79 & 0.72 & 0.78 \\
010,100 & 000,001,011,101,110,111 & 0.64 & 0.94 & 0.85 & 0.79 & 0.86 \\
010,101 & 000,001,011,100,110,111 & 0.42 & 0.83 & 0.74 & 0.76 & 0.78 \\
010,110 & 000,001,011,100,101,111 & 0.56 & 0.82 & 0.90 & 0.79 & 0.84 \\
010,111 & 000,001,011,100,101,110 & 0.48 & 0.84 & 0.74 & 0.76 & 0.78 \\
011,100 & 000,001,010,101,110,111 & 0.69 & 0.93 & 0.81 & 0.77 & 0.84 \\
011,101 & 000,001,010,100,110,111 & 0.52 & 0.81 & 0.68 & 0.76 & 0.75 \\
011,110 & 000,001,010,100,101,111 & 0.56 & 0.83 & 0.80 & 0.77 & 0.80 \\
011,111 & 000,001,010,100,101,110 & 0.59 & 0.80 & 0.68 & 0.78 & 0.75 \\
100,101 & 000,001,010,011,110,111 & 0.74 & 0.93 & 0.79 & 0.71 & 0.81 \\
100,110 & 000,001,010,011,101,111 & 0.77 & 0.99 & 0.66 & 0.78 & 0.81 \\
100,111 & 000,001,010,011,101,110 & 0.72 & 0.94 & 0.78 & 0.75 & 0.82 \\
101,110 & 000,001,010,011,100,111 & 0.52 & 0.83 & 0.75 & 0.75 & 0.78 \\
101,111 & 000,001,010,011,100,110 & 0.60 & 0.80 & 0.65 & 0.75 & 0.73 \\
110,111 & 000,001,010,011,100,101 & 0.61 & 0.84 & 0.77 & 0.71 & 0.77 \\
000,001,010 & 011,100,101,110,111 & 0.76 & 0.99 & 0.69 & 0.74 & 0.81 \\
000,001,011 & 010,100,101,110,111 & 0.74 & 0.95 & 0.75 & 0.76 & 0.82 \\
000,001,100 & 010,011,101,110,111 & 0.65 & 0.82 & 0.99 & 0.73 & 0.85 \\
000,001,101 & 010,011,100,110,111 & 0.73 & 0.93 & 0.82 & 0.73 & 0.83 \\
000,001,110 & 010,011,100,101,111 & 0.62 & 0.95 & 0.91 & 0.74 & 0.87 \\
000,001,111 & 010,011,100,101,110 & 0.72 & 0.93 & 0.80 & 0.74 & 0.82 \\
000,010,011 & 001,100,101,110,111 & 0.72 & 0.99 & 0.69 & 0.82 & 0.83 \\
000,010,100 & 001,011,101,110,111 & 0.65 & 0.83 & 0.78 & 0.94 & 0.85 \\
000,010,101 & 001,011,100,110,111 & 0.69 & 0.98 & 0.67 & 0.91 & 0.85 \\
000,010,110 & 001,011,100,101,111 & 0.66 & 0.92 & 0.78 & 0.94 & 0.88 \\
000,010,111 & 001,011,100,101,110 & 0.70 & 0.98 & 0.67 & 0.91 & 0.85 \\
000,011,100 & 001,010,101,110,111 & 0.59 & 0.81 & 0.94 & 0.89 & 0.88 \\
000,011,101 & 001,010,100,110,111 & 0.64 & 0.93 & 0.77 & 0.88 & 0.86 \\
000,011,110 & 001,010,100,101,111 & 0.57 & 0.93 & 0.87 & 0.88 & 0.89 \\
000,011,111 & 001,010,100,101,110 & 0.65 & 0.93 & 0.78 & 0.90 & 0.87 \\
000,100,101 & 001,010,011,110,111 & 0.67 & 0.81 & 0.94 & 0.84 & 0.86 \\
000,100,110 & 001,010,011,101,111 & 0.65 & 0.84 & 0.75 & 0.93 & 0.84 \\
000,100,111 & 001,010,011,101,110 & 0.62 & 0.81 & 0.94 & 0.90 & 0.88 \\
000,101,110 & 001,010,011,100,111 & 0.57 & 0.94 & 0.85 & 0.89 & 0.89 \\
000,101,111 & 001,010,011,100,110 & 0.67 & 0.92 & 0.75 & 0.89 & 0.85 \\
000,110,111 & 001,010,011,100,101 & 0.59 & 0.94 & 0.88 & 0.84 & 0.89 \\
001,010,011 & 000,100,101,110,111 & 0.64 & 0.86 & 0.76 & 0.85 & 0.82 \\
001,010,100 & 000,011,101,110,111 & 0.57 & 0.94 & 0.89 & 0.87 & 0.90 \\
001,010,101 & 000,011,100,110,111 & 0.51 & 0.83 & 0.77 & 0.87 & 0.82 \\
001,010,110 & 000,011,100,101,111 & 0.54 & 0.82 & 0.94 & 0.90 & 0.89 \\
001,010,111 & 000,011,100,101,110 & 0.53 & 0.83 & 0.76 & 0.88 & 0.82 \\
001,011,100 & 000,010,101,110,111 & 0.65 & 0.93 & 0.79 & 0.92 & 0.88 \\
001,011,101 & 000,010,100,110,111 & 0.67 & 0.81 & 0.67 & 0.94 & 0.81 \\
001,011,110 & 000,010,100,101,111 & 0.60 & 0.84 & 0.77 & 0.93 & 0.85 \\
001,011,111 & 000,010,100,101,110 & 0.69 & 0.80 & 0.65 & 0.96 & 0.80 \\
001,100,101 & 000,010,011,110,111 & 0.67 & 0.93 & 0.84 & 0.81 & 0.86 \\
001,100,110 & 000,010,011,101,111 & 0.69 & 0.99 & 0.69 & 0.89 & 0.86 \\
001,100,111 & 000,010,011,101,110 & 0.64 & 0.93 & 0.82 & 0.87 & 0.87 \\
001,101,110 & 000,010,011,100,111 & 0.53 & 0.83 & 0.80 & 0.88 & 0.84 \\
001,101,111 & 000,010,011,100,110 & 0.63 & 0.79 & 0.67 & 0.90 & 0.79 \\
001,110,111 & 000,010,011,100,101 & 0.59 & 0.84 & 0.81 & 0.83 & 0.83 \\
010,011,100 & 000,001,101,110,111 & 0.63 & 0.95 & 0.90 & 0.75 & 0.87 \\
010,011,101 & 000,001,100,110,111 & 0.54 & 0.85 & 0.80 & 0.74 & 0.80 \\
010,011,110 & 000,001,100,101,111 & 0.61 & 0.81 & 0.97 & 0.74 & 0.84 \\
010,011,111 & 000,001,100,101,110 & 0.57 & 0.84 & 0.80 & 0.75 & 0.80 \\
010,100,101 & 000,001,011,110,111 & 0.65 & 0.94 & 0.86 & 0.76 & 0.85 \\
010,100,110 & 000,001,011,101,111 & 0.73 & 0.93 & 0.74 & 0.84 & 0.84 \\
010,100,111 & 000,001,011,101,110 & 0.62 & 0.94 & 0.86 & 0.80 & 0.87 \\
010,101,110 & 000,001,011,100,111 & 0.54 & 0.82 & 0.91 & 0.81 & 0.85 \\
010,101,111 & 000,001,011,100,110 & 0.48 & 0.84 & 0.73 & 0.80 & 0.79 \\
010,110,111 & 000,001,011,100,101 & 0.60 & 0.82 & 0.93 & 0.75 & 0.83 \\
011,100,101 & 000,001,010,110,111 & 0.69 & 0.94 & 0.82 & 0.74 & 0.83 \\
011,100,110 & 000,001,010,101,111 & 0.73 & 0.98 & 0.68 & 0.80 & 0.82 \\
011,100,111 & 000,001,010,101,110 & 0.67 & 0.93 & 0.82 & 0.80 & 0.85 \\
011,101,110 & 000,001,010,100,111 & 0.54 & 0.83 & 0.80 & 0.78 & 0.80 \\
011,101,111 & 000,001,010,100,110 & 0.60 & 0.80 & 0.68 & 0.81 & 0.76 \\
011,110,111 & 000,001,010,100,101 & 0.61 & 0.83 & 0.83 & 0.74 & 0.80 \\
100,101,110 & 000,001,010,011,111 & 0.78 & 0.99 & 0.66 & 0.75 & 0.80 \\
100,101,111 & 000,001,010,011,110 & 0.73 & 0.94 & 0.78 & 0.73 & 0.82 \\
100,110,111 & 000,001,010,011,101 & 0.78 & 1.00 & 0.68 & 0.74 & 0.81 \\
101,110,111 & 000,001,010,011,100 & 0.61 & 0.85 & 0.77 & 0.73 & 0.78 \\
000,001,010,011 & 100,101,110,111 & 0.79 & 1.00 & 0.66 & 0.72 & 0.79 \\
000,001,010,100 & 011,101,110,111 & 0.62 & 0.84 & 0.81 & 0.76 & 0.80 \\
000,001,010,101 & 011,100,110,111 & 0.74 & 0.98 & 0.71 & 0.76 & 0.82 \\
000,001,010,110 & 011,100,101,111 & 0.68 & 0.94 & 0.81 & 0.78 & 0.84 \\
000,001,010,111 & 011,100,101,110 & 0.74 & 0.98 & 0.69 & 0.76 & 0.81 \\
000,001,011,100 & 010,101,110,111 & 0.60 & 0.82 & 0.92 & 0.78 & 0.84 \\
000,001,011,101 & 010,100,110,111 & 0.73 & 0.94 & 0.76 & 0.79 & 0.83 \\
000,001,011,110 & 010,100,101,111 & 0.64 & 0.95 & 0.85 & 0.77 & 0.86 \\
000,001,011,111 & 010,100,101,110 & 0.73 & 0.93 & 0.75 & 0.80 & 0.83 \\
000,001,100,101 & 010,011,110,111 & 0.65 & 0.81 & 1.00 & 0.71 & 0.84 \\
000,001,100,110 & 010,011,101,111 & 0.57 & 0.85 & 0.80 & 0.78 & 0.81 \\
000,001,100,111 & 010,011,101,110 & 0.60 & 0.81 & 0.98 & 0.76 & 0.85 \\
000,001,101,110 & 010,011,100,111 & 0.61 & 0.94 & 0.92 & 0.77 & 0.88 \\
000,001,101,111 & 010,011,100,110 & 0.72 & 0.93 & 0.80 & 0.78 & 0.84 \\
000,001,110,111 & 010,011,100,101 & 0.63 & 0.95 & 0.92 & 0.72 & 0.86 \\
000,010,011,100 & 001,101,110,111 & 0.60 & 0.84 & 0.80 & 0.86 & 0.83 \\
000,010,011,101 & 001,100,110,111 & 0.70 & 0.99 & 0.70 & 0.85 & 0.85 \\
000,010,011,110 & 001,100,101,111 & 0.66 & 0.93 & 0.81 & 0.85 & 0.86 \\
000,010,011,111 & 001,100,101,110 & 0.70 & 0.98 & 0.70 & 0.86 & 0.85 \\
000,010,100,101 & 001,011,110,111 & 0.65 & 0.83 & 0.79 & 0.91 & 0.84 \\
000,010,100,110 & 001,011,101,111 & 0.70 & 0.80 & 0.65 & 1.00 & 0.82 \\
000,010,100,111 & 001,011,101,110 & 0.59 & 0.83 & 0.78 & 0.96 & 0.86 \\
000,010,101,110 & 001,011,100,111 & 0.64 & 0.93 & 0.79 & 0.97 & 0.90 \\
000,010,101,111 & 001,011,100,110 & 0.68 & 0.98 & 0.66 & 0.95 & 0.86 \\
000,010,110,111 & 001,011,100,101 & 0.66 & 0.93 & 0.80 & 0.90 & 0.88 \\
000,011,100,101 & 001,010,110,111 & 0.58 & 0.82 & 0.95 & 0.86 & 0.88 \\
000,011,100,110 & 001,010,101,111 & 0.54 & 0.83 & 0.75 & 0.92 & 0.83 \\
000,011,100,111 & 001,010,101,110 & 0.54 & 0.81 & 0.95 & 0.92 & 0.89 \\
000,011,101,110 & 001,010,100,111 & 0.55 & 0.94 & 0.88 & 0.90 & 0.91 \\
000,011,101,111 & 001,010,100,110 & 0.64 & 0.93 & 0.77 & 0.94 & 0.88 \\
000,011,110,111 & 001,010,100,101 & 0.57 & 0.94 & 0.90 & 0.85 & 0.90 \\
000,100,101,110 & 001,010,011,111 & 0.65 & 0.84 & 0.75 & 0.89 & 0.83 \\
000,100,101,111 & 001,010,011,110 & 0.63 & 0.82 & 0.93 & 0.87 & 0.87 \\
000,100,110,111 & 001,010,011,101 & 0.63 & 0.85 & 0.76 & 0.88 & 0.83 \\
000,101,110,111 & 001,010,011,100 & 0.59 & 0.95 & 0.86 & 0.87 & 0.89 \\
    \hline
    \caption{A full summary of the results of the procedure described in Section~\ref{subsec:full_dec_process}. $\bm{S_1}$ and $\bm{S_2}$: the decomposition; \textbf{1\textsuperscript{st} Level}: inter-annotator agreement ($\kappa$) on $C_{S_1}$; \textbf{Complication}, \textbf{Resolution} and \textbf{Success}: inter-annotator agreement ($\kappa$) on each element given an agreement on $C_{S_1}$; \textbf{Average}: second-level inter-annotator agreement, averaged over the three narrative elements. All the statistics were averaged over the two annotator pairs.}
    \label{tab:full_results}
\end{longtable}
}
\twocolumn

\end{document}